\relax
\documentclass[letterpaper]{article} 
\usepackage{aaai21}  
\usepackage{times}  
\usepackage{helvet} 
\usepackage{courier}  
\usepackage[hyphens]{url}  
\usepackage{graphicx} 
\usepackage{natbib}  
\usepackage{caption} 
\usepackage{amsmath}
\usepackage{amssymb}
\usepackage{bm}
\usepackage{subcaption}
\usepackage[switch]{lineno}

\title{\texttt{TRACE}: Early Detection of Chronic Kidney Disease Onset with Transformer-Enhanced Feature Embedding}
\author {
	Yu Wang \textsuperscript{\rm 1},
	Ziqiao Guan \textsuperscript{\rm 1}, 
	Wei Hou \textsuperscript{\rm 2}, 
	Fusheng Wang \textsuperscript{\rm 1,3}\\
} 
\affiliations {
	\textsuperscript{\rm 1} Department of Computer Science, Stony Brook University \\
	\textsuperscript{\rm 2} Department of Family, Population \& Preventive Medicine, Stony Brook University \\
	\textsuperscript{\rm 3} Department of Biomedical Informatics, Stony Brook University
}

\begin{document}
	
	
	\maketitle

	\begin{abstract}
		
		Chronic kidney disease (CKD) has a poor prognosis due to excessive risk factors and comorbidities associated with it. The early detection of CKD faces challenges of \emph{insufficient medical histories of positive patients} and \emph{complicated risk factors}. In this paper, we propose the \texttt{TRACE} (\underline{T}ransformer-\underline{R}NN \underline{A}utoencoder-enhanced \underline{C}KD D\underline{e}tector) framework, an end-to-end prediction model using patients' medical history data, to deal with these challenges. \texttt{TRACE} presents a comprehensive medical history representation with a novel key component: a Transformer-RNN autoencoder. The autoencoder jointly learns a medical concept embedding via Transformer for each hospital visit, and a latent representation which summarizes a patient's medical history across all the visits. We compared \texttt{TRACE} with multiple state-of-the-art methods on a dataset derived from real-world medical records. Our model has achieved 0.5708 AUPRC with a 2.31\% relative improvement over the best-performing method. We also validated the clinical meaning of the learned embeddings through visualizations and a case study, showing the potential of \texttt{TRACE} to serve as a general disease prediction model.  
		
	\end{abstract}

	\section{Introduction}
	
	Chronic kidney disease (CKD) is a general term for many heterogeneous diseases that irreversibly alter kidney structure or cause a chronic reduction in kidney function~\cite{levey2012chronic}. It is defined by the presence of kidney damage, or decreased kidney function, or both, for a minimum of three months. Diagnosis of CKD is often made after accidental findings from screening laboratory tests, or when the symptom already becomes severe~\cite{webster2017chronic}. According to the US Centers for Disease Control and Prevention, approximately 15\% of US adults have CKD, but most people may not feel ill or notice any symptoms until CKD is advanced~\cite{ckd}. Two major causes of CKD are hypertension and diabetes. Kidney failure is the most serious outcome of CKD and severe conditions can only be treated by dialysis and transplantation, which are indications of end-stage kidney disease. CKD is a well-known risk factor for cardiovascular disease and all-cause mortality~\cite{abboud2010stage,tonelli2006chronic}. However, less than 2\% of CKD patients finally require renal replacement therapy, because many of them die from cardiovascular causes before end-stage kidney disease can occur~\cite{keith2004longitudinal}. 
	
	CKD has a poor prognosis due to excessive risk factors and comorbidities associated with it~\cite{kronenberg2009emerging}. The early detection of CKD, for health benefits, is an even more challenging task. For the purposes of early detection and control, a promising direction is to regularly monitor risk factors of CKD for high-risk patients. It is also worth trying to focus the screening for CKD on younger, healthier populations, although it is less likely to detect CKD in such cohorts~\cite{tonelli2006chronic}. In short, early detection of CKD faces challenges of \emph{insufficient medical histories of positive patients} and \emph{complicated risk factors} from a data standpoint. This calls for more effective machine learning prediction models to address these issues.
	
	To address the first challenge, we need a prediction model that can \emph{better extract knowledge from the insufficient medical histories}. In recent years, deep learning has emerged as a powerful tool to gain insight into EHR data~\cite{xiao2018opportunities,ching2018opportunities,miotto2018deep,shickel2017deep}. Previous studies predicted heart failure onset with recurrent neural networks (RNNs)~\cite{choi2017using} and the reverse time attention mechanism~\cite{choi2016retain}. These models proposed different ideas for sequence modeling, but they did not perform very well in our experiments. Apart from sequence modeling, we also need a better feature representation for the early detection of CKD onset. This ties in with the second challenge -- complicated risk factors. 
	
	Given how extensive the risk factors are when assessing a patient's likelihood of having CKD, it is critical to \emph{learn an embedding to measure the latent similarity between these risk factors and CKD}. An embedding maps discrete medical concepts to a continuous latent space and summarizes the interactions between medical concepts. Several models can learn embeddings of medical concepts~\cite{choi2016multi,choi2018mime,choi2020learning}. However, these models have yet to make the most of the sequential nature of EHR data because they only learned medical concept embeddings for individual hospital visits. We can use a sequence aggregator, such as RNN or 1-D convolutional neural networks (CNNs), to link a sequence of visits and encode them to a patient-level representation. 
	
	Motivated by the challenges and the previous work, we propose the \texttt{TRACE} (\underline{T}ransformer-\underline{R}NN \underline{A}utoencoder-enhanced \underline{C}KD D\underline{e}tector) framework, which combines ideas of both RNN autoencoder and Transformer~\cite{vaswani2017attention,choi2020learning}. \texttt{TRACE} jointly learns the Transformer-encoded hidden structure of individual hospital visits while predicting CKD onset with RNN. In this work, we used Cerner Health Facts~\cite{deshazo2015comparison}, a large database derived from EHR systems across the US, as our data source. The database includes comprehensive patient-level details of diagnoses, procedures, medications, and laboratory tests. The EHR data are structured data without clinical notes and images. Our features include both medical codes and non-medical code information. We summarize our contributions as follows:
	\begin{itemize}
		\item To the best of our knowledge, this is the first work that builds advanced sequential deep learning models to predict CKD onset. 
		\item We propose a Transformer-RNN autoencoder architecture. This autoencoder jointly learns a medical concept embedding via Transformer for individual hospital visits, as well as a patient embedding via an RNN encoder-decoder structure which summarizes the entire medical history of this patient. 
		\item We adopt two pre-training processes to capture proper latent representations for patients' conditions and hospital visit histories respectively. We also demonstrated that \texttt{TRACE} successfully alleviated the aforementioned challenges by incorporating the pre-trained latent representations.
	\end{itemize}

	\section{Related Work}

	\subsection{Deep Learning in Healthcare Domain}
	
	Deep learning algorithms have become popular approaches for modeling disease progression~\cite{choi2016doctor,ma2017dipole,choi2016retain,zachary2016learning}, patient characterization~\cite{baytas2017patient,che2015deep}, and generating synthetic EHR data for research purposes~\cite{choi2017generating}.
	
	The most common application of modeling disease progression is predicting disease outcomes. Deep neural networks have very limited power when learning disease trajectories from scratch, sometimes it is necessary to incorporate prior medical knowledge~\cite{ma2018risk,pham2017predicting} or supplement EHR data with inherent hierarchical structure of medical ontologies~\cite{choi2017gram}. Missing value is also a challenge is modeling EHR data. ~\cite{che2018recurrent} has demonstrated that RNNs are able to capture the long-term dependency in time series and improve prediction performance. 
	
	Deep learning models anticipate a large volume of data to achieve satisfactory results, which usually exceeds the capacity of most healthcare facilities. A straightforward solution is to combine EHR data from multiple sources, but data harmonization is a labor-intensive process.~\cite{rajkomar2018scalable} recently proposed a representation of EHRs based on the Fast Healthcare Interoperability Resources (FHIR) format for deep learning models without site-specific data harmonization. When working on an imbalanced dataset with insufficient positive samples, data augmentation techniques can benefit training. CONAN~\cite{cui2020conan} incorporated generative adversarial networks (GANs) to create candidate positive and negative samples in rare disease detection. Pre-training and transfer learning~\cite{bengio2012deep,dauphin2012unsupervised} can also help to solve this problem. G-BERT~\cite{shang2019pretraining} used the pre-trained hospital visit representations for downstream predictive tasks.~\cite{rios2019neural} trained a CNN on a large global database with biomedical abstracts, and transferred the learned knowledge to predict diagnosis codes for one medical center. 
	
	\subsection{Representation Learning for Medical Concepts}
	
	Representation learning algorithms in healthcare domain are mainly borrowed from natural language processing (NLP). The general idea is to encode discrete medical concepts (e.g., medical codes) to one-hot vectors~\cite{bengio2013representation} and then apply Word2Vec algorithms~\cite{tomas2013efficient} to learn embeddings. 
	
	For example, Med2Vec~\cite{choi2016multi} utilized skip-gram~\cite{mikolov2013distributed} to learn intra-visit medical code co-occurrences as well as inter-visit sequential information. The generic skip-gram model is based on the assumption that a word can play different roles at different positions in a sentence. However, this assumption doesn't hold for medical codes given their unordered nature. When we adopt NLP algorithms to model medical concepts, we typically apply the algorithms to the feature dimension instead of the temporal dimension, and the order in which these medical concepts occur is ignored.
	
	MiME~\cite{choi2018mime} leveraged the inherent structure of medical codes to learn a multilevel embedding of EHR data, but this model required the EHR data to contain complete structure information between diagnoses and treatments. Recently, GCT~\cite{choi2020learning} has been proposed to solve this problem. GCT learned the graphical structure of EHR data during training and proved that Transformer is a suitable model to learn such structure. Our work was motivated by GCT to use Transformer to encode hospital visits.

	\section{Method}

	\subsection{Problem Statement}
	
	We formulate this problem as a binary prediction task. Given a patient whose medical history is in the form of a sequence of hospital visits $\mathcal{P} = \{\mathcal{V}_1, \mathcal{V}_2, \cdots, \mathcal{V}_T\}$ in chronological order, where $T \in \mathbb{N}$ is the total number of visits that the patient has. Each visit $\mathcal{V}_{t}\ (t \in \{1,2,\cdots,T\})$ consists of a list of medical codes, clinical observations and other information related to this patient. We want to predict whether the patient will be diagnosed with CKD for the first time in the following visit $\mathcal{V}_{T+1}$.

	\subsection{Vector Representations of EHRs}
	
	\begin{figure}[t]
		\centering
		\includegraphics[width=\columnwidth]{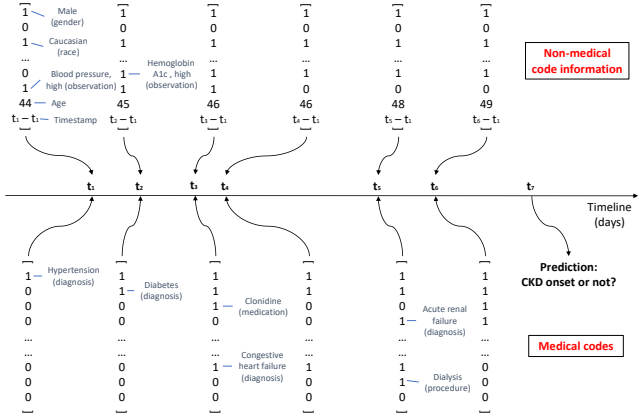}
		\caption{Vector representation of a patient's EHR data. This patient had seven hospital visits at time $\{t_1, t_2, \cdots, t_7\}$. We used the first six visits $\{t_1, \cdots, t_6\}$ to construct input vectors to the model and predicted whether there would be a CKD onset at the seventh visit.} 
		\label{fig:vector-representation}
	\end{figure}
	
	Figure~\ref{fig:vector-representation} illustrates the vector representations of a patient's hospital visit history $\mathcal{P}$, which contains both medical codes and non-medical code information. For simplicity, all notations and algorithms in this paper are presented for a single patient unless otherwise specified.

	\subsubsection{Medical Code Representations}
	
	Medical codes include diagnosis codes, procedure codes, and medication codes. Medical codes are \emph{the primary features} for our prediction task. We denote the set of medical codes in our EHR data by $\mathcal{C} = \{c_1, c_2, \cdots, c_{|\mathcal{C}|}\}$ with size $|\mathcal{C}|$. All medical codes that occur at a hospital visit $\mathcal{V}_t$ are represented by a multi-hot vector $\bm{x}_t \in \{0,1\}^{|\mathcal{C}|}$ where the $i$-th element is 1 if $c_i \in \mathcal{V}_t$. 
	
	\subsubsection{Non-medical Code Information}
	
	Besides medical codes, we also included the patient's observations (e.g., lab tests, vital signs, etc.), age, race, gender, and the timestamp of visit $\mathcal{V}_t$. Observations are \emph{the secondary features} in our dataset. Let $\bm{d}_t$ denote the vector representation of the non-medical code information, which is a concatenation of multi-hot vector and numeric values. We will provide details of theses non-medical code features in the ``Dataset" section.

	\subsection{Model Architecture of \texttt{TRACE}}
	
\begin{figure*}[t]
		\centering
		\includegraphics[width=0.95\textwidth]{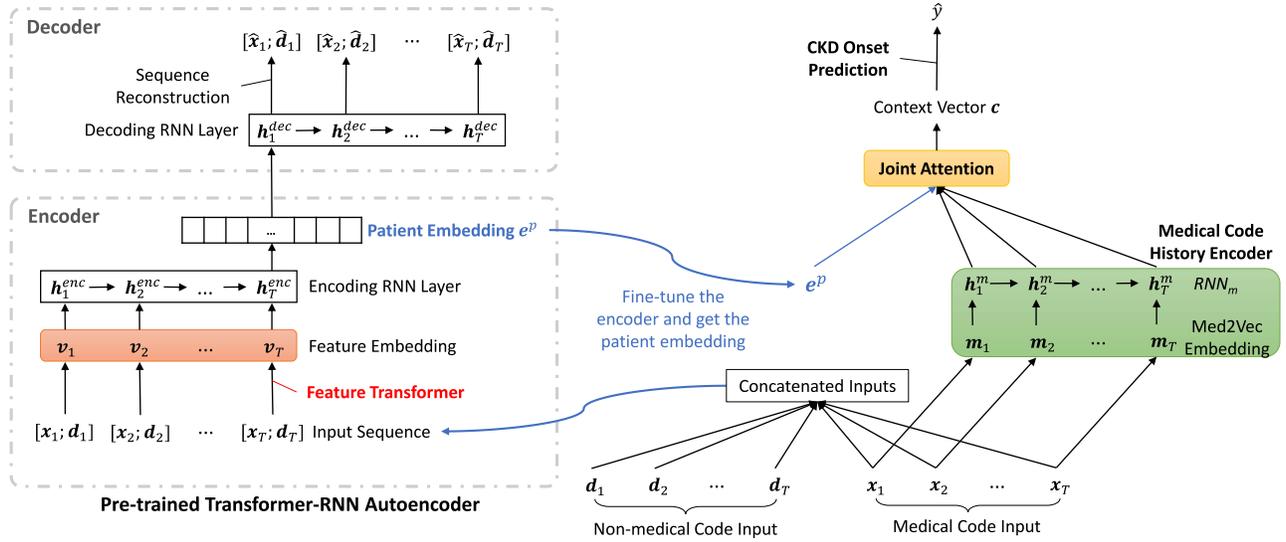}
		\caption{End-to-end structure of \texttt{TRACE}. The model ingests a patient's medical history in the form of two sequences of vectors $\mathcal{X} = \{\bm{x}_1, \bm{x}_2, \cdots, \bm{x}_T\}$ and $\mathcal{D} = \{\bm{d}_1, \bm{d}_2, \cdots, \bm{d}_T\}$, next propagates them to the encoder module of a pre-trained Transformer-RNN autoencoder for a patient embedding, then combines the patient embedding with the patient's medical code history to compute joint attention, and finally outputs a probability score $\hat{y}$.}
		\label{fig:model-architecture}
	\end{figure*}

	In this section, we describe \texttt{TRACE} in detail, with the following components: a patient embedding from a pre-trained Transformer-RNN autoencoder, a medical code history encoder, and a joint attention module. The overall architecture is illustrated in Figure~\ref{fig:model-architecture}.
	
	\subsubsection{Patient Embedding via Transformer-RNN Autoencoder}
	
	We pre-train a Transformer-RNN autoencoder as a joint feature extractor for both feature embedding and patient embedding. This design adopts a Transformer~\cite{choi2020learning} for computing self-correlation of all features at individual hospital visit level, and a subsequent RNN autoencoder for learning a patient representation by reconstructing the input sequence in an unsupervised fashion.
	
	The autoencoder ingests a sequence of hospital visits in the form of $\{\bm{x}'_t\}_{t=1}^{T}$ where $\bm{x}'_t=[\bm{x}_t; \bm{d}_t] \in \mathbb{R}^{n \times 1}$ and $n$ is the total number of features. The sequence $\{\bm{x}'_t\}_{t=1}^{T}$ then runs through an encode-decoder structure to reconstruct itself. 
	\begin{itemize}
		\item \textbf{Encoder}. We first map each $\bm{x}'_t\ (t\in \{1,2,\cdots,T\})$ to a latent space with a learnable embedding matrix $\bm{W}_x$ by $\bm{Z}_t = \bm{W}_x \odot \bm{x}'_t \in \mathbb{R}^{n \times d_z}$. This extends the raw input vector $\bm{x}'_t$ to a \emph{vector array} for the Transformer to process. Since Transformer has a quadratic time and space complexity and is expensive to compute, we downsize the feature dimension from $n$ to $\tilde{n}\ (\tilde{n} \ll n)$  via a linear transform $\tilde{\bm{Z}}_t = \tilde{\bm{W}_z} \bm{Z}_t \in \mathbb{R}^{\tilde{n} \times d_z}$. This improves scalability by reducing the complexity to $\mathcal{O}(\tilde{n}^2)$. 
		
		We learn an embedding for the downsized feature space using a Transformer with one encoder block and a single attention head as follows,
		\begin{equation}
			\bm{X}_t = \mathrm{Transformer}(\tilde{\bm{Z}}_t),
		\end{equation}
		where $\bm{X}_t \in \mathbb{R}^{\tilde{n} \times d_\mathrm{emb}}$. Positional encoding is removed from our framework, since the features are not ordered.
		
		To aggregate the embedding of all features occurred at time $t$, we average-pool $\bm{X}_t$ by the downsized feature dimension to obtain a single embedding vector $\bm{v}_t  \in \mathbb{R}^{d_{\mathrm{emb}}}$. Eventually, we feed the sequence $\{\bm{v}_t\}_{t=1}^T$ in an encoding RNN layer to encode the entire input sequence to a single-vector patient embedding $\bm{e}^p$.
		\item \textbf{Decoder}. We propagate $\bm{e}^p$ to a decoding RNN layer to obtain a decoded sequence $\{\bm{h}_t^{\mathrm{dec}}\}_{t=1}^T$, then reconstruct the input sequence from $\{\bm{h}_t^{\mathrm{dec}}\}_{t=1}^T$. Specifically, we add separate fully-connected layers on top of $\{\bm{h}_t^{\mathrm{dec}}\}_{t=1}^T$ to reconstruct different types of features, and obtain a reconstructed sequence $\{\hat{\bm{x}}'_t\}_{t=1}^T$. The loss between the input sequence $\{\bm{x}'_t\}_{t=1}^T$ and the reconstructed sequence $\{\hat{\bm{x}}'_t\}_{t=1}^T$ is the sum of multiple losses. For multi-hot medical codes and observations, we use softmax classifiers with cross entropy loss. For race and gender, we apply sigmoid classifiers with cross entropy loss. For age and timestamp, we apply linear transform and minimize mean squared errors. 
		\item \textbf{Patient Embedding for \texttt{TRACE}}. To obtain a patient embedding for our end-to-end prediction task, we feed the input sequence $\{[\bm{x}_t; \bm{d}_t]\}_{t=1}^T$ in the pre-trained encoder and compute the embedding vector $\bm{e}^p$ for this patient (Figure~\ref{fig:model-architecture}).
	\end{itemize}
	
	We also summarize the detailed pre-training process as an algorithm in the appendix.

	\subsubsection{Medical Code History Encoder}
	
	Considering that medical codes summarize other non-medical code features to some extent, we separately encode the patient's medical code history for information gain in our model. 
	
	We first map the discrete medical code inputs $\bm{x}_t$ to a continuous latent embedding space $\bm{m}_t$ as follows,
	\begin{equation}
		\bm{m}_t = \bm{W}_{m} \bm{x}_t,
		\label{eq:medcode-embedding-med2vec}
	\end{equation}
	where $\bm{W}_{m} \in \mathbb{R}^{|\mathcal{C}| \times d_m}$ is a word embedding lookup table pre-trained via Med2Vec~\cite{choi2016multi}, and $d_m$ is the size of the embedding vector. To encode a sequence of medical codes, we then apply an RNN layer on top of the medical code embedding $\bm{m}_t$ by
	\begin{equation}
		\bm{h}_t^m = \mathrm{RNN}_m (\bm{h}_{t-1}^m, \bm{m}_t)
		\label{eq:medcode-rnn}
	\end{equation} 
	where $\bm{h}_t^m$ is the hidden state of the RNN layer at time $t$.
	
	\subsubsection{Joint Attention}
	
	We want to further have the patient embedding interact with the medical code history. Specifically, we compute interactions between $\bm{e}^p$ and each $\bm{h}_t^m\ (t\in \{1,2,\cdots,T\})$ as follows,
	\begin{align}
		\bm{g}_t &= [\bm{e}^p; \bm{h}_t^m], \\
		\mathrm{score}_t &= \bm{u}^\top \tanh (\bm{W}_g \bm{g}_t + \bm{b}_g), \\
		\alpha_t &= \frac{\exp(\mathrm{score}_t)}{\sum_{t=1}^{T}\exp(\mathrm{score}_t)},
		\label{eq:attention}
	\end{align}
	where $\bm{u}$ is a learable weight, and $\alpha_t$ is the attention weight assigned to visit $\mathcal{V}_t$. Then we obtain a context vector $\bm{c}$ for this patient by
	\begin{equation}
		\bm{c} = [\bm{e}^p; \sum_{t=1}^{T} \alpha_t \bm{g}_t].
	\end{equation}
	
	\subsubsection{CKD Onset Prediction}
	
	We use the context vector $\bm{c}$ to predict the binary label $y \in \{0, 1\}$ as follows,
	\begin{equation}
		\hat{y} = \sigma(\bm{w}_y^\top \bm{c} + b_y),
	\end{equation}
	where $\hat{y}$ is the predicted probability score for this patient. The training objective is to use the predicted score $\hat{y}$ and the true label $y$ to minimize the following binary cross entropy loss:
	\begin{equation}
		\mathcal{L} = - \frac{1}{N} \sum_{j=1}^{N} (y_j \log\hat{y}_j + (1-y_j)\log(1-\hat{y}_j)),
		\label{eq:log-loss}
	\end{equation}
	where $N$ is the total number of patients in our training set.

	\section{Experiments}
	
	\subsection{Dataset}
	
	We collected our experimental dataset from two healthcare systems in Cerner Health Facts, each healthcare system comprises multiple healthcare facilities. This is a \emph{case-control study} where negative patients were downsampled through a statistical analysis, such that our model was trained to distinguish positive and negative patients who were similar in terms of age, race and gender.

	\subsubsection{Features and Data Preprocessing}
	
	We extracted a patient's diagnosis codes, procedure codes, medication codes, observations, age, race, gender, and admission date for features of each hospital visit. Statistics of the features are available in Table~\ref{tab:dataset}. 
	
	The raw diagnosis codes, procedure codes and medication codes in Cerner Health Facts are respectively International Classification of Diseases (ICD), Current Procedural Terminology (CPT) and generic drug names. We grouped ICD diagnosis codes by the Clinical Classifications Software (CCS) to obtain higher-level diagnosis codes for experiments, which reduced the number of diagnosis codes from over 69,000 to 275. CKD diagnoses were identified by the CCS codes and were excluded from the feature set. We did not consider hospital visits without any diagnosis codes documented and removed medical codes that appeared in less than 50 hospital visit records. 
	
	Apart from medical codes, we also included observations, age, race, gender, and admission date to represent a hospital visit. Observations, race and gender were categorical features encoded to a multi-hot vector (Figure~\ref{fig:vector-representation}) for a hospital visit. There were 1,261 distinct observations in our feature set. For the admission date of a patient's hospital visit, we converted it to a numeric timestamp by calculating the duration in days from the patient's first visit to this visit. We took the logarithm of numeric features for model inputs. 
	
	\subsubsection{Selection of Cases and Controls} 
	
	We excluded hospital visits made by non-adult patients because we aimed at predicting CKD onset for adults only. The case/control selection criteria are as follows.
	\begin{itemize}
		\item \emph{Cases (positives)} were patients who had at least one hospital visit prior to CKD onset. The CKD onset was the positive class label for our prediction task.
		\item \emph{Controls (negatives)} were non-CKD patients who had at least two hospital visits in our dataset. Controls were identified for each case using the propensity score-matching based on logistic regression~\cite{rosenbaum1983central} and the greedy algorithm~\cite{bergstralh1995computerized}. Matching variables include age, gender and race. Class labels of control patients came from their latest hospital visit records and were negative labels.
	\end{itemize}
	
	Six controls were selected for each case to match the prevalence of CKD in US adults (i.e., $1/7 \approx 14.29\%$). Eventually, we extracted a total of 147,791 patients for experiments. The dataset was further split into training, validation and test sets in a 75/10/15 ratio. The case/control ratio in each of the training, validation and test sets was the same as the disease prevalence rate in the entire experimental dataset. Table~\ref{tab:dataset} provides details of the study cohort. Since 90\% of patients in the dataset had less than 30 hospital visits, we only kept up to 30 most recent hospital visits per patient to improve scalability. 
	
	\begin{table}[t]
		\centering
		\resizebox{.95\columnwidth}{!}{
			\begin{tabular}{ccc}
				\hline
				& Experiments & Med2Vec \\
				\hline
				Total \# of patients & 147,791 & 1,155,450 \\
				\# of cases (positives) & 21,113 & N/A \\
				\# of controls (negatives) & 126,678 & N/A \\
				\# of patients for training & 110,842 & N/A \\
				\# of patients for validation & 14,778 & N/A \\
				\# of patients for testing & 22,171 & N/A \\
				\hline
				Total \# of medical codes & 1,679 & 3,884 \\
				\# of diagnosis codes & 275 & 278 \\
				\# of procedure codes & 662 & 2,449 \\
				\# of medication codes & 742 & 1,157 \\
				\hline
				Total \# of observations & 1,261 & N/A \\
				Total \# of races \& genders & 10 & N/A \\
				\hline
			\end{tabular}
		}
		\caption{Statistics of datasets for our experiments and the pre-training using the Med2Vec model.}
		\label{tab:dataset}
	\end{table}

	\subsection{Pre-training for Medical Concepts}
	
	There were two types of medical concepts in our dataset: medical codes and observations. We performed two pre-training processes to get proper embeddings for them.
	
	\subsubsection{Independent Pre-training}
	
	We pre-trained embedding weights for \emph{medical codes}\footnote{We also trained Med2Vec to obtain an embedding for observations, but got very poor results.} using Med2Vec. The dataset for this pre-training task was extracted from 10 healthcare facilities in Cerner Health Facts, not including the two healthcare systems for our experiments. We trained the Med2Vec model on 1,155,450 patients with 15,115,251 hospital visits and obtained pre-trained embedding weights for 3,884 distinct medical codes (Table~\ref{tab:dataset}). In our prediction task, we treated medical codes outside the independent pre-training as out-of-vocabulary (OOV) tokens and initialized embedding weights for OOV medical codes with zeros.
	
	\subsubsection{Transformer-Encoded Embedding}
	
	We pre-trained the Transformer-RNN autoencoder on our training set to encode \emph{all features}, where only age, race, gender, and timestamp were not medical concepts (Table~\ref{tab:dataset}). This means that the Transformer-encoded feature embedding was a good latent representation of medical concepts. The pre-trained feature embedding was built into \texttt{TRACE} as part of the encoder module for fine-tuning (Figure~\ref{fig:model-architecture}).

	\subsection{Baseline Models}
	
	For comparison, we implemented the following models with $\bm{x}'_t = [\bm{x}_t; \bm{d}_t]$ as the input vector for a hospital visit.
	\begin{itemize}
		\item\textbf{Logistic regression (LR)}. We counted the occurrences of each medical code and each observation for a patient, all the other features were determined by the patient's last hospital visit in the inputs. A LR model was trained on the resulting vectors.
		\item\textbf{Multi-layer perceptron (MLP)}. We used the same approach to construct model inputs as the LR model, but added a fully-connected layer with relu activation between the input layer and the output layer.
		\item\textbf{RNN} and \textbf{BiRNN}. We used a fully-connected layer with relu activation to encode inputs and then propagated the resulting vectors to a forward/bidirectional RNN layer. Logistic regression was applied to the last hidden state of the RNN layer to predict CKD onset. 
		\item\textbf{RETAIN}~\cite{choi2016retain}. RETAIN model was designed to predicts heart failure onset using backward RNN and two levels of attention weights. We used the same architecture as the RNN baseline, but replaced the RNN layer with the RETAIN module. 
		\item\textbf{Dipole}~\cite{ma2017dipole}. Dipole model predicts multiple disease outcomes via a bidirectional RNN layer and three different attention mechanisms. We used the same structure as the RNN baseline, but replaced the RNN layer with the Dipole module and trained it using each of the three attention mechanisms, i.e., Dipole\textsubscript{l}, Dipole\textsubscript{g} and Dipole\textsubscript{c}. 
		\item\textbf{1-D CNN}. A modification of AlexNet~\cite{krizhevsky2012imagenet}. We replaced all 2-D convolutional layers with 1-D convolutional layers, which served as a sequence aggregator of individual hospital visits. We computed the mean of AlexNet's outputs across the temporal dimension and applied logistic regression on top of it to generate predictions. The inputs were encoded in the same way as the RNN baseline.
	\end{itemize}

	\subsection{Evaluation Metrics}
	
	We measured the model performance on our test set by area under the precision-recall curve (AUPRC). AUPRC can effectively evaluate the fraction of true positives among positive predictions~\cite{saito2015precision}, thus it is an appropriate metric when evaluating binary classifiers on imbalanced datasets like ours. In addition to AUPRC, we also calculated negative log likelihood by Eq.~\ref{eq:log-loss} to measure the model loss on the test set.

	\subsection{Implementation Details}
	
	We implemented all models and calculated all evaluation metrics using TensorFlow 2.2.0~\cite{tensorflow2015-whitepaper}. For Med2Vec, we used the code provided by the authors\footnote{https://github.com/mp2893/med2vec}. The dimension of the downsized feature space was $\tilde{n}=100$. The sizes of all embedding vectors and hidden layers were 128. The dropout rate for the feed-forward layer of Transformer was 0.5. We used the Adadelta optimizer~\cite{zeiler2012adadelta} and set the learning rate as 1.0 to match the exact form in the paper. We trained each model for 50 epochs with 100 patients per batch. All experiments were run on a 16GB NVIDIA Tesla V100 PCIe GPU.

	\subsection{Results}
	
	\begin{table}[t]
		\centering
		\resizebox{.99\columnwidth}{!}{
			\begin{tabular}{cccc}
				\hline
				Category & Model & AUPRC & Neg log likelihood \\
				\hline
				Non- & LR & 0.4527 & 0.3453 \\
				sequence & MLP & 0.5359 & 0.3067 \\
				\hline
				CNN & 1-D CNN & 0.5475 & 0.3017 \\
				\hline
				& RNN & 0.5574 & 0.2978 \\
				& BiRNN & 0.5510 & 0.2986 \\
				RNN & RETAIN & 0.5505 & 0.2986 \\
				& Dipole\textsubscript{l} & 0.5563 & 0.2969 \\
				& Dipole\textsubscript{g} & 0.5579 & 0.2994 \\
				& Dipole\textsubscript{c} & 0.5515 & 0.2962 \\
				\hline
				Ours & \texttt{TRACE} & \textbf{0.5708} & \textbf{0.2929} \\
				\hline
			\end{tabular}
		}
		\caption{Prediction performance of different models.}
		\label{tab:auprc}
	\end{table}
	
	\begin{table}[t]
		\centering
		\begin{tabular}{ccc}
			\hline
			Model & AUPRC & Neg log likelihood \\
			\hline
			RACE\_base & 0.5649 & 0.2955 \\
			\hline
			RACE & 0.5631 & 0.2938 \\
			\hline
			\texttt{TRACE}\_base & 0.5696 & 0.2937 \\
			\hline
			\texttt{TRACE} & \textbf{0.5708} & \textbf{0.2929} \\
			\hline
		\end{tabular}
		\caption{Ablation study of \texttt{TRACE}.}
		\label{tab:ablation-study}
	\end{table}

	\begin{figure}[t]
		\centering
		\begin{subfigure}[b]{0.495\columnwidth}
			\centering
			\includegraphics[width=\columnwidth]{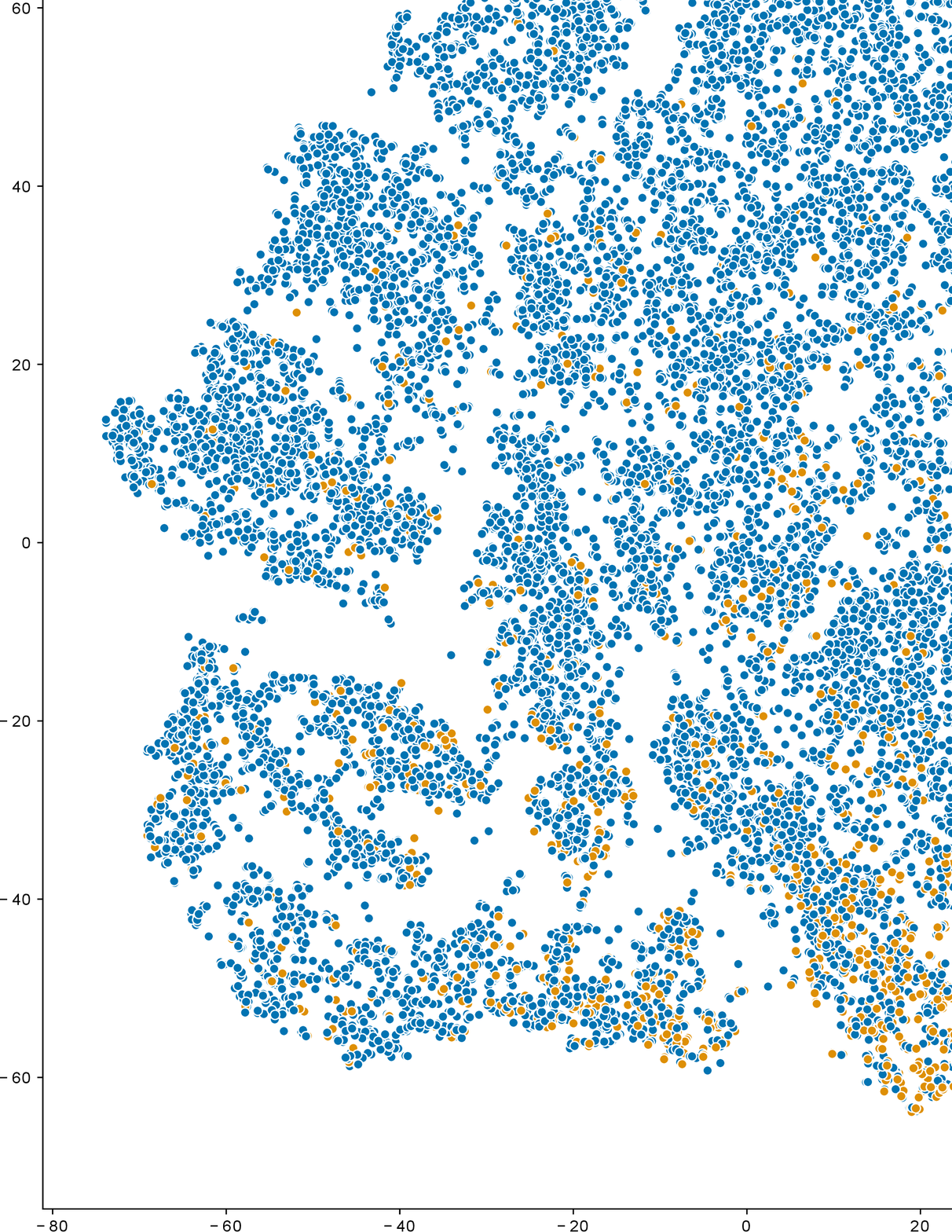}
			\caption{RNN.}
			\label{fig:relu-rnn-patient-embedding}
		\end{subfigure}
		\hfill
		\begin{subfigure}[b]{0.495\columnwidth}
			\centering
			\includegraphics[width=\columnwidth]{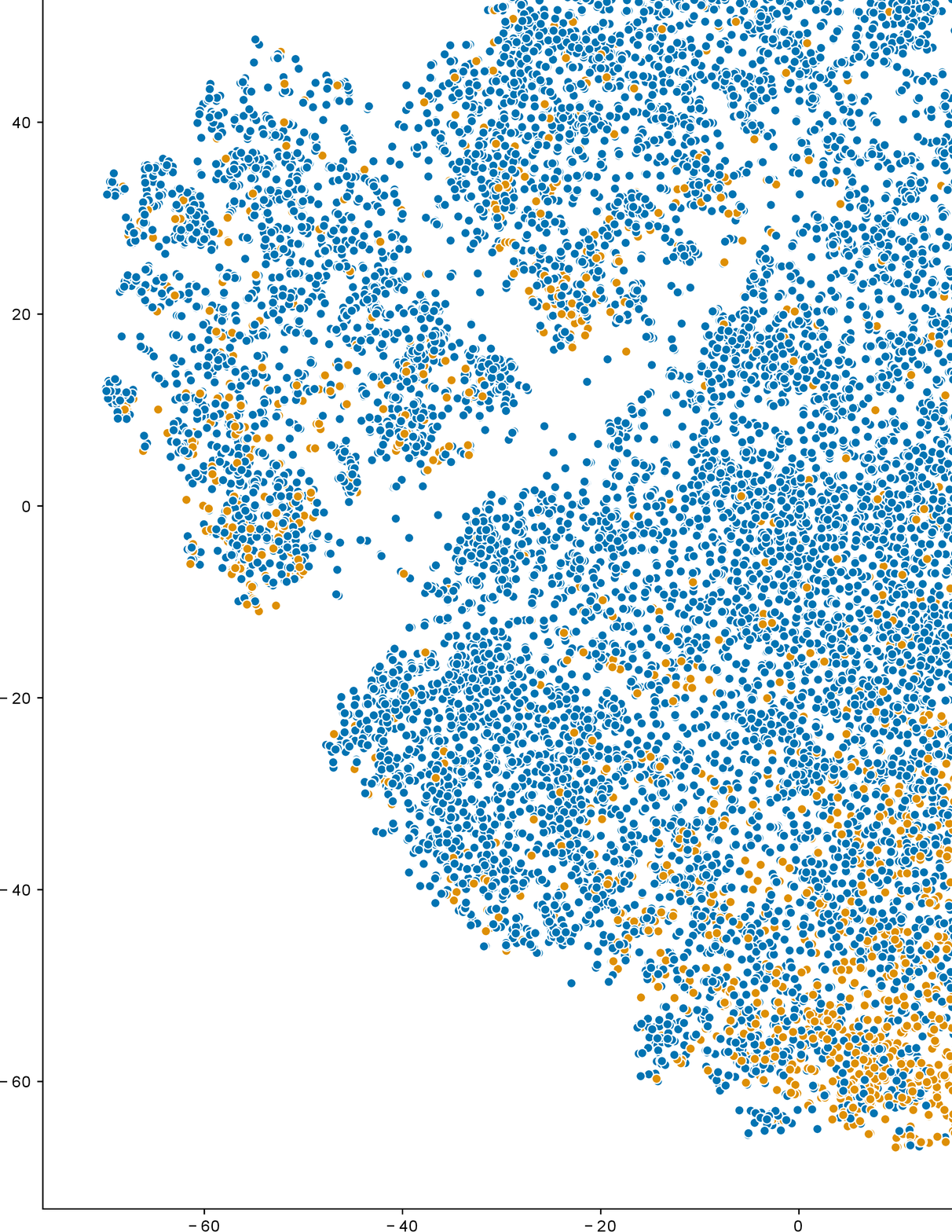}
			\caption{\texttt{TRACE}.}
			\label{fig:trace-patient-embedding}
		\end{subfigure}
		\caption{Patient embeddings learned by different models for the test set. Dimension reduced via t-SNE. Orange dot: positive patient, blue dot: negative patient. \texttt{TRACE} yielded a clearer boundary between positive and negative patients, and produced a better clustering of positive patients.}
		\label{fig:patient-embedding}
	\end{figure}
	
	\begin{figure*}[t]
		\centering
		\includegraphics[width=0.99\textwidth]{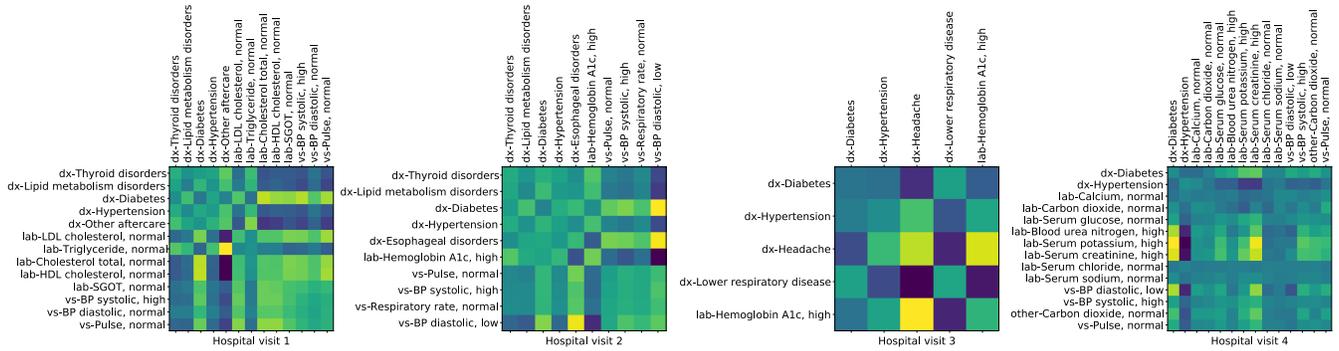}
		\caption{Medical concept attentions produced by \texttt{TRACE} for a CKD patient in the test set. The labels are the medical concepts occurred at each hospital visit. It illustrates how medical concepts on the vertical axis attended to medical concepts on the horizontal axis. ``dx-": diagnosis code (medical code), ``lab-": lab test (observation), ``vs-": vital sign (observation), ``other-": observation other than lab test and vital sign, ``BP": blood pressure. This patient got a 0.9198 prediction score with four hospital visits as inputs.}
		\label{fig:trace-attention}
	\end{figure*}

	\subsubsection{Prediction Performance}
	
	We evaluated our model and all baselines on the test set. Table~\ref{tab:auprc} shows the AUPRC and negative log likelihood scores of the test set. In general, sequential models (RNNs and 1-D CNN) outperformed non-sequential models (LR and MLP). This is because sequential models were more capable of capturing the underlying patterns in disease trajectories, while non-sequential models only learned from aggregated information of medical histories. In the real-world clinical practice, doctors need to carefully review a patient's medical histories and monitor the patient for a long time to decide whether a chronic disease like CKD can be diagnosed. Occasional occurrences of some symptoms related to CKD are insufficient to support the doctor's decision.
	
	It is noteworthy that all RNN-based baselines demonstrated comparable performance in terms of both metrics, and increased model complexity failed to surpass the simplest RNN model. This seems to indicate that training RNN models from scratch is not suitable for our task. Both RETAIN and Dipole computed attention scores with the outputs of RNN layers. The attention mechanism assigns a score to each hospital visit using the sequential information learned from scratch, which is fine when detecting diseases in their original tasks. However, CKD is quite different because its excessive risk factors could be intertwined. It is hard to determine whether a patient has CKD simply by the existence of several risk factors without extensive prior knowledge.
	
	\texttt{TRACE} successfully alleviated this problem by introducing a pre-trained Transformer-RNN autoencoder. The autoencoder produced a good patient embedding which compressed information in the entire input sequence. With this patient embedding as the prior knowledge, our end-to-end prediction model was able to better discover the correlation between CKD diagnoses and past medical records. \texttt{TRACE} achieved a 2.31\% gain in AUPRC compared with the best-performing baseline (i.e., Dipole\textsubscript{g}).

	\subsubsection{Ablation Study}
	
	To understand how each major model component contributed to the overall prediction performance, we compared \texttt{TRACE} with its several variants.
	\begin{itemize}
		\item\textbf{\texttt{TRACE}\_base}. This is \texttt{TRACE} without the medical code history encoder and the joint attention. We directly used the fine-tuned patient embedding to get predictions.
		\item\textbf{RACE}. This is \texttt{TRACE} without Transformer-encoded feature embedding. In the RNN autoencoder, we got a feature embedding through a fully-connected layer with relu activation instead.
		\item\textbf{RACE\_base}. This is RACE without the medical code history encoder and the joint attention. The fine-tuned patient embedding were directly used for getting predictions.
	\end{itemize}
	
	We trained the three variants with the same set of hyperparameters as \texttt{TRACE}. We note the AUPRC scores for the analyses here (Table~\ref{tab:ablation-study}). Overall, pre-trained RNN autoencoders provided richer patient-level information than raw input features. Even the worst-performing model in Table~\ref{tab:ablation-study} (i.e., RACE) achieved a 0.93\% relative improvement in AUPRC over the best-performing baseline in Table~\ref{tab:auprc} (i.e., Dipole\textsubscript{g}). Evidently, Transformer has demonstrated its superiority over pure fully-connected layers in encoding medical concepts (RACE vs. \texttt{TRACE} and RACE\_base vs. \texttt{TRACE}\_base). As we expected, there was slight information gain after adding medical code histories and the joint attention, but the strength was limited (RACE\_base vs. RACE and \texttt{TRACE}\_base vs. \texttt{TRACE}).

	\subsubsection{Patient Embedding Visualization}
	
	Figure~\ref{fig:patient-embedding} plots patient embeddings produced by \texttt{TRACE} and the RNN baseline respectively. We used the t-SNE~\cite{maaten2008visualizing} algorithm for dimensionality reduction. Obviously, \texttt{TRACE} learned a clearer boundary between positive and negative patients, as well as a better clustering of positive patients. Given that this is a case-control study, \texttt{TRACE} met our expectation to better distinguish cases and controls who were similar in terms of age, race and gender. We also provide an illustration of pre-trained patient embeddings in the appendix. Transformer-RNN autoencoder was able to produce a more gathered patient embedding than the generic RNN autoencoder, showing the strength of Transformer in encoding features.

	\subsubsection{Attention Visualization and Case Study}
	
	We visualize the attention behavior of Transformer in the course of CKD onset prediction. Since \texttt{TRACE} computed self-attention for the downsized feature space $\tilde{\bm{Z}_t}$, we need to back-propagate to the original feature space by $ \bm{Z}_t = \tilde{\bm{W}_z}^\top \tilde{\bm{Z}}_t$ to get desired attention weights. To improve readability of the visualization, we randomly selected a CKD patient from the test set, who had four hospital visits and at most 20 medical concepts per visit.
	
	Figure~\ref{fig:trace-attention} illustrates the attention behavior of medical concepts occurred at each hospital visit for the selected patient, which also shows the patient's disease trajectory. This patient had hypertension and diabetes -- two major causes of CKD that usually intertwine with other risk factors of CKD. No remarkable attention behavior was present at the first hospital visit. At the second visit, we noticed that esophageal disorders, diabetes and low diastolic blood pressure were mutually attended. At the third visit, the high hemoglobin A1c level and headache attended to each other, indicating a poor blood sugar control and a higher risk of diabetes complications. Eventually, at the fourth visit, the patient had a group of CKD risk factors tested, such as blood urea nitrogen, serum potassium and serum creatinine. The abnormal test results all attended to diabetes, which suggested the correlation between diabetes and CKD. The high serum potassium level also attended to the high serum creatinine level. Moreover, this patient got a true positive prediction with a 0.9198 prediction score.
	
	We also visualize the pre-trained attention behavior for the same patient in the appendix, which was produced by the pre-trained Transformer-RNN autoencoder. The pre-trained attention behavior is similar to the one fine-tuned by \texttt{TRACE}, but some CKD risk factors stood out after fine-tuning.

	\section{Conclusion}
	
	In this work, we proposed the \texttt{TRACE} framework, a novel end-to-end prediction model that incorporated a pre-trained Transformer-RNN autoencoder for early detection of CKD onset. It is hard to predict CKD onset by training a model from scratch due to the excessive risk factors and insufficient medical histories of positive patients. \texttt{TRACE} alleviated this problem by introducing prior knowledge learned by the autoencoder. Experimental analyses showed that \texttt{TRACE} outperformed all baselines and its several variants in predicting CKD onset. We also validated the clinical meaning of the learned embeddings through visualizations and a case study, which demonstrated the potential of \texttt{TRACE} to be generalized to other disease prediction tasks. In the future, we plan to combine data augmentation techniques like GAN to better address the data insufficiency. We will also adopt more advanced NLP algorithms to train embeddings for patients and features.


	\bibliography{references}

\end{document}